\documentclass[11pt]{article}

\usepackage[preprint]{acl}

\usepackage{times}
\usepackage{latexsym}

\usepackage[T1]{fontenc}

\usepackage[utf8]{inputenc}

\usepackage{microtype}

\usepackage{inconsolata}

\usepackage{graphicx}

\usepackage{amsmath}
\usepackage{amssymb}
\usepackage{mathtools}
\usepackage{amsthm}
\usepackage{enumitem}
\usepackage{booktabs}
\usepackage{multirow}
\usepackage[table]{xcolor}
\usepackage{booktabs,tabularx}
\usepackage{listings}
\usepackage{enumitem}
\usepackage{tcolorbox}
\tcbuselibrary{skins}
\usepackage{booktabs,multirow,array,tabularx}
\usepackage{xcolor,colortbl}

\definecolor{resultgray}{gray}{0.95}

\providecommand{\resultNA}{\textemdash}

\lstdefinestyle{skillscopeEditorJSON}{
  basicstyle=\small\ttfamily,
  breaklines=true,
  breakatwhitespace=false,
  columns=fullflexible,
  keepspaces=true,
  showstringspaces=false,
  aboveskip=3pt,
  belowskip=3pt
}

\newtcolorbox{skillscopeEditorPrompt}[1]{
  enhanced,
  width=\linewidth,
  colback=black!3,
  colframe=black!75,
  colbacktitle=black!75,
  coltitle=white,
  fonttitle=\normalsize\bfseries,
  title={#1},
  boxrule=0.7pt,
  arc=1.5mm,
  left=8pt,
  right=8pt,
  top=6pt,
  bottom=6pt,
  fontupper=\small
}

\title{Before the Next Rollout: Diagnosing Skill Utility Shifts from Policy Updates}

\author{
    \begin{tabular}{c}
        \textbf{Yifan Wang\textsuperscript{1,2,*}},
        \textbf{Zhenyi Gai\textsuperscript{2,*}},
        \textbf{Yiwen Zhao\textsuperscript{1}},
        \textbf{Mingxuan Jiang\textsuperscript{1}},
        \textbf{Keyang Chen\textsuperscript{1}},
                              \\[0.4em]
        \textbf{Zeping Li\textsuperscript{1}},
        \textbf{Hongfeng Chai\textsuperscript{1}},
        \textbf{Yixin Cao\textsuperscript{1,$\dagger$}}
        \\[0.8em]
        {\small
        \textsuperscript{1}Fudan University
        \qquad
        \textsuperscript{2}Pyra
        }
        \\[0.4em]
        {\small
        \textsuperscript{*}Equal Contribution
        \qquad
        \textsuperscript{$\dagger$}Correspondence to:
        \texttt{yxcao@fudan.edu.cn}
        }
    \end{tabular}
}

\begin{document}
\maketitle

\begin{abstract}
Skill-evolution methods typically analyze execution trajectories to guide agent skill refinement.
However, policy updates can change the behavioral effects and marginal utility of an unchanged skill. Attribution based on earlier trajectories may therefore no longer reflect the skill's contribution under the updated policy.
Moreover, trajectory failures can arise from multiple factors, making individual skill contributions difficult to isolate.
We evaluate whether policy-update signals contain predictive information about changes in skill marginal utility. 
At fixed contexts from pre-update training rollouts, we measure how policy updates change the contrast in token preferences between skill and control conditions. Combining these interaction shifts with advantage estimates yields reward-directed readouts for diagnosing utility changes. The readouts reuse existing training rollouts and require only forward passes for single-step action scoring, without additional post-update rollout evaluation. 
Our experiments show that policy updates can alter skill marginal utility and that policy-update readouts carry diagnostic information for identifying skills with declining utility. Integrating readout-guided prioritization into skill evolution improves downstream performance while reducing policy and editor token consumption.\footnote{Code is available at \url{https://anonymous.4open.science/r/SkillSense/}}



\end{abstract}

\section{Introduction}
\begin{figure}[tp]
  \centering
  \includegraphics[width=\columnwidth]{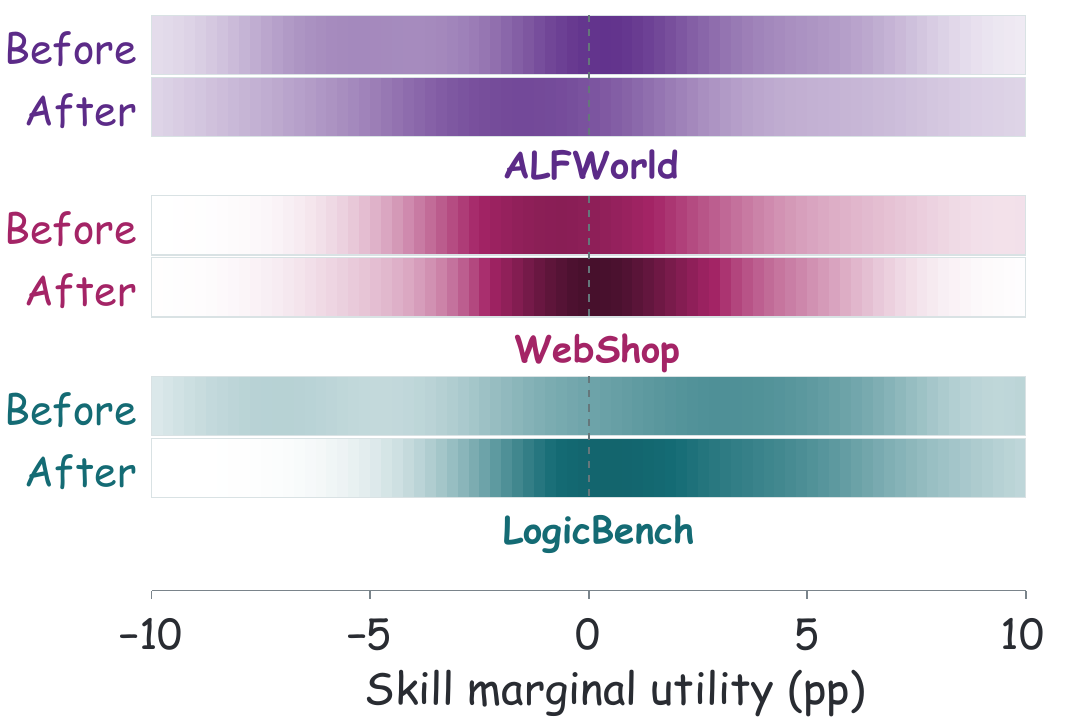}
  \caption{
    \textbf{Policy updates alter skill marginal utility.}
    Distributions before and after policy updates, with skill content fixed.
    Darker shades indicate higher concentrations of skills
    at the corresponding utility values.
}
  \label{fig:policy-utility-shifts}
\end{figure}

Large language model (LLM) agents have demonstrated strong capabilities in tackling complex tasks through multi-step reasoning, tool use, and interaction with external environments~\cite{jin2025searchr1trainingllmsreason, zheng-etal-2025-deepresearcher}. Reusable skills augment these agents by organizing task knowledge and procedures into external guidance that can be applied to subsequent tasks~\cite{liang2026skillnetcreateevaluateconnect}. To improve their adaptability and transferability, recent work iteratively evaluates and revises skills using accumulated interaction experience~\cite{xia2026skillrlevolvingagentsrecursive, tang2026wikiskillcompilingagentexperience}.

\begin{figure*}[tp]
  \centering
  \includegraphics[width=0.90\textwidth]{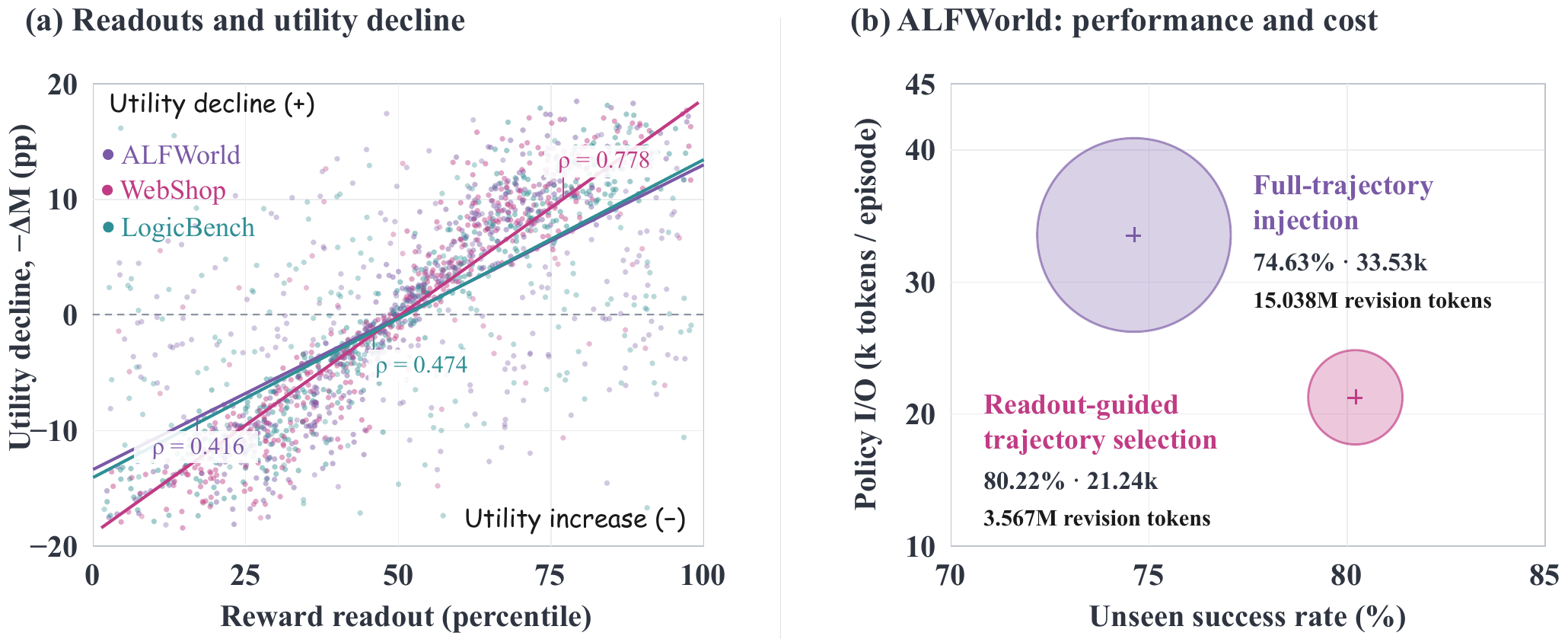}
\caption{
    \textbf{Diagnostic value and downstream benefits of policy-update readouts.}
    (a) Association between reward-directed readout percentiles and
    measured utility declines across three benchmarks.
    Positive $-\Delta M$ indicates declining skill utility;
    $\rho$ denotes Spearman correlation.
    (b) ALFWorld unseen success rate versus policy token consumption
    under full-failure and readout-guided revision.
    Bubble size indicates cumulative editor token usage.
}
  \label{fig:readout-evidence}
\end{figure*}

As skill libraries grow, identifying which skills remain useful and which warrant reassessment or revision becomes increasingly important. 
Here, skill marginal utility measures the difference in expected task success between using a skill and a matched control that removes its task-specific guidance.
Existing skill-evolution methods typically rely on execution trajectories and the resulting feedback~\cite{zheng2025skillweaverwebagentsselfimprove, ni2026trace2skilldistilltrajectorylocallessons}. However, skill utility is policy-dependent: even when a skill remains unchanged, policy updates can alter its behavioral effects and marginal
contribution (Fig.~\ref{fig:policy-utility-shifts}). Consequently, evidence from earlier trajectories may no longer accurately reflect skill utility under the updated policy~\cite{he2026reskillreconcilingskillcreation}, while repeated evaluation through fresh rollouts adds to the cost of ongoing maintenance. Moreover, failures in agent trajectories can arise from multiple factors, making individual skill contributions difficult to isolate from execution results alone~\cite{ding2026agentskillevaluationevolution}. This raises a key question: Can we diagnose changes in skill utility from policy updates without additional environment rollouts?


To address this challenge, we investigate the predictive value of policy-update signals for changes in skill marginal utility. Holding skill content and decision contexts fixed, we compare how policy updates change the skill--control contrast in token preferences. This captures changes in the behavioral influence of unchanged skill guidance, while advantage estimates from the training batch help assess whether these shifts align with reward-favored behavior.
These readouts reuse existing training trajectories and require only forward passes for single-step action scoring, without additional post-update rollouts for diagnosis. By prioritizing skills with potentially declining utility, they focus subsequent revision on a smaller set of candidates, while associated failure trajectories guide the actual edits. 
Our experiments show that reward-directed readouts are positively
associated with measured utility declines across three benchmarks
(Fig.~\ref{fig:readout-evidence}(a)), supporting the diagnostic
value of policy-update signals.
On ALFWorld, readout-guided revision further improves unseen
task success while reducing policy and editor token consumption
(Fig.~\ref{fig:readout-evidence}(b)).

Our main contributions include:
\begin{itemize}[leftmargin=1.2em]
    \item We show that policy updates can alter the marginal utility of unchanged skills, and that changes in policy responses provide predictive information about these utility shifts before additional post-update rollout evaluation.

   \item We construct reward-directed readouts by combining policy-response
changes with advantage estimates from existing training batches.
These readouts provide diagnostic information about skill utility
shifts without requiring additional post-update rollouts.

   \item We integrate readout-guided prioritization into a controlled
skill-revision loop, demonstrating improved downstream performance
and reduced policy and editor token consumption relative to
full-failure-based revision under a shared editor and matched
revision budgets.
\end{itemize}

\section{Related Work}
\textbf{Experience-driven skill evolution.} Recent work enables LLM agents to accumulate reusable guidance through feedback-driven prompt optimization, context adaptation, and experience-based memory~\cite{agrawal2026gepareflectivepromptevolution, zhang2026agenticcontextengineeringevolving, ouyang2026reasoningbankscalingagentselfevolving}. Research on agent skills further explores reusable procedure acquisition, transferable representations, and the organization of skill libraries~\cite{zheng2025skillweaverwebagentsselfimprove, yu2026polyskilllearninggeneralizableskills, liang2026skillnetcreateevaluateconnect}. Building on these foundations, trajectory-based approaches distill execution experience into persistent skills, preserve accumulated lessons, and stabilize subsequent revisions using historical feedback~\cite{ni2026trace2skilldistilltrajectorylocallessons, tang2026wikiskillcompilingagentexperience, li2026skilladamstableefficientskill}. These developments support iterative improvement of external guidance through interaction. Another line of work integrates skill evolution with reinforcement learning, allowing skill libraries and executing policies to adapt jointly~\cite{xia2026skillrlevolvingagentsrecursive}. This coupling also makes the relevance of past evidence time-dependent: ReSkill discounts historical feedback as the policy evolves~\cite{he2026reskillreconcilingskillcreation}, while PATS~\cite{shi2026patspolicyawaretrainingscaffolding} reuses training-rollout evidence to adapt scaffolds for subsequent learning. Although these methods address aspects of policy adaptation and feedback reuse, historical feedback reflects the behavior of the policy that generated it and may not reliably capture skill utility after an update, potentially weakening subsequent maintenance decisions.

\textbf{Skill utility evaluation and prediction.} Skill evaluation increasingly measures the incremental benefit of external guidance through comparisons with and without skills, revealing heterogeneous gains across tasks and agent configurations~\cite{li2026skillsbenchbenchmarkingagentskills, han2026sweskillsbenchagentskillsactually}. Related approaches incorporate contribution estimates into training and lifecycle management: SLIM~\cite{shen2026dynamicskilllifecyclemanagement} uses leave-one-skill-out evaluations to inform retention, while UCOB~\cite{tu2026ucoblearningutilizeevolve} compares skill-conditioned and unconditioned returns at matched task–state anchors to guide distillation. These studies provide foundations for measuring marginal contributions. Predictive approaches additionally seek to reduce the execution cost of assessing skills. Cross-task transferability can be estimated from skill and task descriptions~\cite{feng2026breakdownpasson}, while COBRA-Skills~\cite{lu2026cobraskillscontextualbanditguidedevolution} learns reward predictors from historical evaluations to allocate budgets among candidate skills under a frozen model. Our work explores how policy-update signals relate to utility shifts in unchanged skills and reuses these signals for low-overhead diagnosis, without training a separate utility predictor or collecting additional post-update environment rollouts.

\section{Methodology}
\begin{figure*}[t]
  \centering
  \includegraphics[width=1\textwidth]{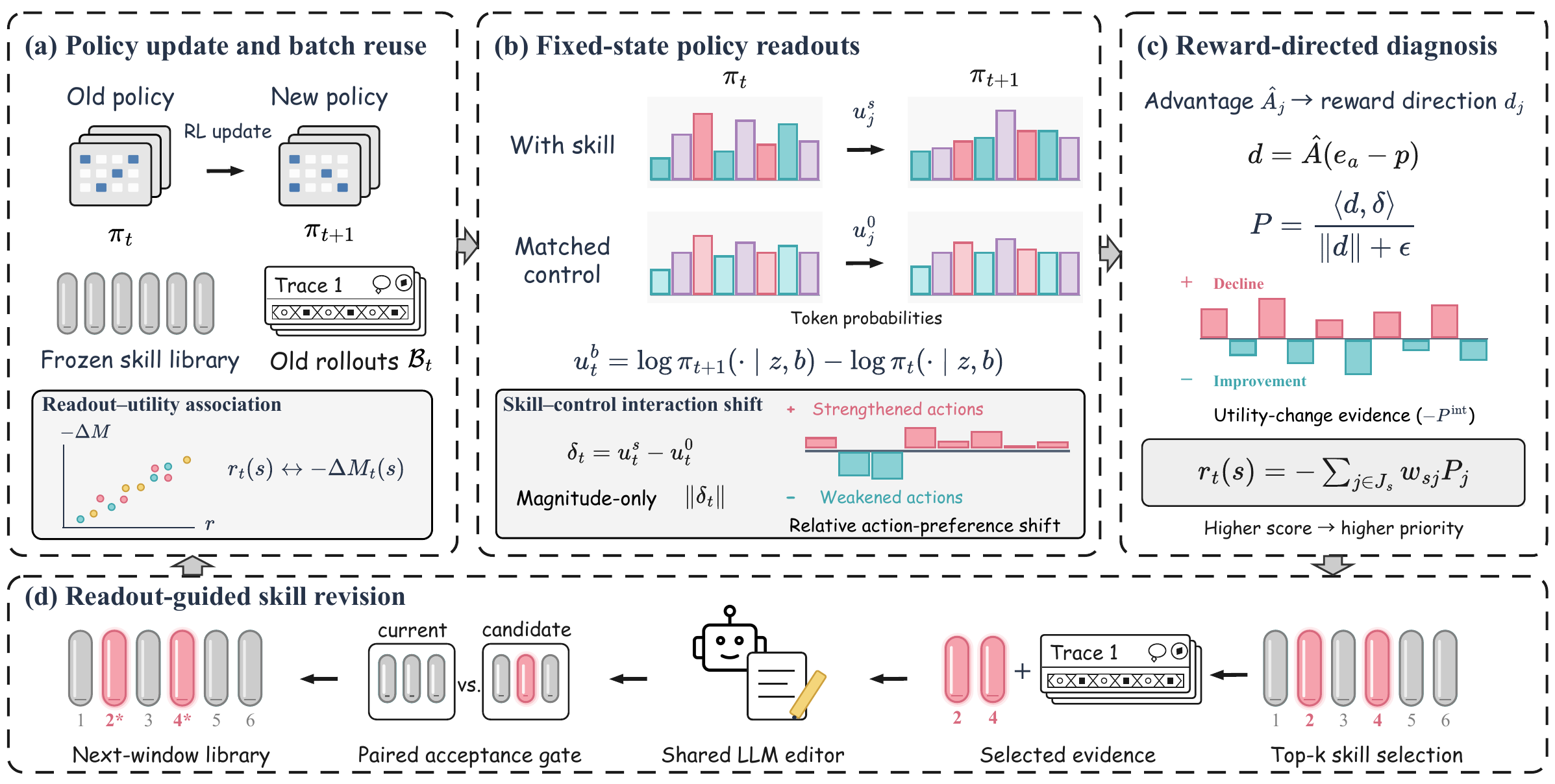}
    \caption{
    \textbf{Overview of our method.}
    With the skill library fixed within each update window, we reuse pre-update
    rollouts to compare action preferences under skill and matched-control
    conditions before and after the policy update.
    Combining these fixed-state comparisons with batch advantages yields
    reward-directed readouts that prioritize skills for revision without
    additional diagnostic rollouts.
}
  \label{fig:method-overview}
\end{figure*}

This section presents our approach to diagnosing skill utility shifts and guiding skill revision, as illustrated in Figure~\ref{fig:method-overview}.
We first define skill marginal utility and measure its change across a policy update (Section~\ref{sec:method-setup}).
We then reuse training contexts to measure changes in the skill--control contrast in token preferences (Section~\ref{sec:method-fixed-state}).
Projecting these interaction shifts onto reward directions yields skill-level readouts for prioritizing potential utility declines (Section~\ref{sec:method-reward}).
Finally, we describe how these readouts prioritize skills for reassessment within a skill-revision loop (Section~\ref{sec:method-revision}).

\subsection{Measuring Skill Utility Shifts}
\label{sec:method-setup}
To evaluate whether policy-update readouts predict changes in skill utility, we first define a paired measurement of each skill's marginal contribution before and after an update.
Let $\pi_t$ and $\pi_{t+1}$ denote the policies at the update window $t$. The skill library $\mathcal K$ remains fixed
within the window, and $s\in\mathcal K$ denotes a specific version of a skill.
At each decision state, we compare the target skill with a control that removes its task-specific guidance. 

\paragraph{Reference anchors.}
For each reference trajectory $\tau$ generated by $\pi_t$ that invokes skill $s$, let $\ell_s(\tau)$ denote its first invocation. We define
\begin{equation}
    \zeta_s(\tau)=\bigl(x,h_{\ell_s(\tau)}\bigr),
    \label{eq:method-anchor}
\end{equation}
where $x$ is the task instance and $h_{\ell_s(\tau)}$ includes the current observation and preceding interaction history.
Both policies continue from the same anchors with equal remaining interaction budgets.

\paragraph{Skill marginal utility.}
Following skill--control performance contrasts~\citep{han2026sweskillsbenchagentskillsactually,tu2026ucoblearningutilizeevolve}, let $V_\pi^b(\zeta;s)$ be the expected task success when $\pi$ continues from $\zeta$, with $b=1$ denoting the original skill and $b=0$ its control. 
We define the marginal utility of skill $s$ and its change across the policy update as
\begin{equation}
\begin{gathered}
M_t(s)
=
\mathbb{E}_{\zeta\sim\nu_s}
\left[
V_{\pi_t}^{1}(\zeta;s)
-
V_{\pi_t}^{0}(\zeta;s)
\right],
\\[4pt]
\Delta M_t(s)
=
M_{t+1}(s)-M_t(s),
\end{gathered}
\label{eq:method-utility}
\end{equation}
where $\Delta M_t(s)<0$ indicates that the skill provides less marginal benefit relative to its control after the policy update. We estimate these quantities using repeated paired continuations across both policies and skill conditions. Uncertainty estimation is detailed in Appendix~\ref{app:method-utility}.

\paragraph{Diagnostic objective.}
Let $\mathcal B_t$ be the training rollout batch collected by $\pi_t$. We construct a readout
\begin{equation}
    r_t(s)=F(\pi_t,\pi_{t+1},\mathcal B_t;s),
    \qquad s\in\mathcal K_t^{\mathrm{inv}},
    \label{eq:method-objective}
\end{equation}
where $\mathcal K_t^{\mathrm{inv}}$ contains the skills invoked in $\mathcal B_t$.
Higher scores assign higher reassessment priority, with the goal of identifying skills whose marginal utility declines.
We compare alternative readouts by their association with $-\Delta M(s)$ and their ability to rank declining skills, as evaluated in Section~\ref{Skill-Level Diagnostic Performance.}.

\subsection{Fixed-State Policy Readouts}
\label{sec:method-fixed-state}
We compare the two policy checkpoints on the same recorded contexts from $\mathcal B_t$, collected under $\pi_t$ at the start of the update window.
Let $z$ denote the recorded context at a response position, comprising the task, observation, interaction history, and response prefix.
Holding $z$ fixed, we forward-score under the skill and control conditions to measure changes in token-level action preferences.

\paragraph{Skill-conditioned and control shifts.}
For each token $a$ in the vocabulary $\mathcal V$, we define the skill-conditioned log-probability shift as
\begin{equation}
    u_t^{s}(z,a)
    =
    \log \pi_{t+1}(a\mid z,s)
    -
    \log \pi_t(a\mid z,s),
    \label{eq:method-skill-shift}
\end{equation}
and the corresponding control shift as
\begin{equation}
    u_t^{\varnothing}(z,a)
    =
    \log \pi_{t+1}(a\mid z,\varnothing)
    -
    \log \pi_t(a\mid z,\varnothing).
    \label{eq:method-control-shift}
\end{equation}
Both shifts are evaluated at the same context.

\paragraph{Skill--control interaction shift.}
To measure how the policy update changes the skill--control contrast, we subtract the control shift from the skill-conditioned shift:
\begin{equation}
    \delta_t^s(z,a)
    =
    u_t^s(z,a)-u_t^{\varnothing}(z,a).
    \label{eq:method-interaction}
\end{equation}
Collecting these differences over the vocabulary yields
$\delta_t^s(z)=(\delta_t^s(z,a))_{a\in\mathcal V}$.
This interaction shift measures how the policy update changes the
skill--control contrast in token preferences, isolating skill-conditioned behavioral
changes from policy shifts shared with the control.

\subsection{Reward-Directed Skill Diagnosis}
\label{sec:method-reward}

The interaction shift describes changes in token preferences, but does not determine whether skill utility increases or decreases.
We combine it with the training advantages in $\mathcal B_t$ to construct a reward-directed readout for assessing the direction of skill utility change.

\paragraph{Local reward direction.}
For a recorded token $a$ at context $z$ under skill $s$, let $\hat A$ be its training advantage and $p=\pi_t(\cdot\mid z,s)$ the pre-update token distribution.
We define
\begin{equation}
    d=\hat A(e_a-p),
    \label{eq:method-direction}
\end{equation}
where $e_a$ is the one-hot vector for $a$.
Since $e_a-p$ is the logit gradient of $\log \pi_t(a\mid z,s)$, positive advantage favors the recorded token and negative advantage disfavors it.

\begin{table*}[tp]
    \centering
    \begingroup
    \small
    \setlength{\tabcolsep}{4pt}
    \renewcommand{\arraystretch}{1.12}

    \newcommand{\skillscopeTablePending}{\textcolor{red}{\textemdash}}

    \begin{tabular*}{\textwidth}{
        @{\hspace{0.35em}}
        l
        @{\extracolsep{\fill}}
        cclrcrrc
        @{}
    }
        \toprule

        \multirow{2}{*}{\raisebox{-0.35ex}{\textbf{Benchmark}}}
        & \multirow{2}{*}{\raisebox{-0.35ex}{\textbf{Skills}}}
        & \multirow{2}{*}{\raisebox{-0.35ex}{\textbf{Decline}}}
        & \multirow{2}{*}{\raisebox{-0.35ex}{\textbf{Readout}}}
        & \multicolumn{4}{c}{\textbf{Decline diagnosis}}
        & \multicolumn{1}{c}{\textbf{Magnitude}} \\

        \cmidrule(lr){5-8}
        \cmidrule(lr){9-9}

        & & & &
        AP$\uparrow$
        & $\mathrm{AUROC}_{-/+}\uparrow$
        & {\normalsize$\rho_{\mathrm{dir}}$}$\uparrow$
        & P@5$\uparrow$
        & {\normalsize $\rho_{\mathrm{mag}}$} \\

        \midrule

        \multirow{2}{*}{\textbf{ALFWorld}}
        & \multirow{2}{*}{25/44}
        & \multirow{2}{*}{32.0\%}
        & Magnitude
        & 0.483
        & 0.643
        & $-0.032$
        & 0.200
        & \textbf{0.170} \\

        & & & Reward
        & \textbf{0.567}
        & \textbf{0.833}
        & \textbf{0.416}
        & \textbf{0.400}
        & $-0.174$ \\

        \addlinespace[0.35em]

        \multirow{2}{*}{\textbf{WebShop}}
        & \multirow{2}{*}{42/54}
        & \multirow{2}{*}{50.0\%}
        & Magnitude
        & 0.739
        & 0.600
        & 0.334
        & 0.400
        & \textbf{0.505} \\

        & & & Reward
        & \textbf{0.853}
        & \textbf{0.800}
        & \textbf{0.778}
        & \textbf{0.800}
        & 0.073 \\

        \addlinespace[0.35em]

        \multirow{2}{*}{\textbf{HotpotQA}}
        & \multirow{2}{*}{22/30}
        & \multirow{2}{*}{36.4\%}
        & Magnitude
        & 0.560
        & 0.542
        & 0.080
        & 0.400
        & \textbf{0.390} \\

        & & & Reward
        & \textbf{0.720}
        & \textbf{0.792}
        & \textbf{0.520}
        & \textbf{0.600}
        & $-0.080$ \\

        \addlinespace[0.35em]

        \multirow{2}{*}{\textbf{NQ}}
        & \multirow{2}{*}{18/30}
        & \multirow{2}{*}{44.4\%}
        & Magnitude
        & 0.640
        & 0.554
        & 0.121
        & 0.400
        & \textbf{0.441} \\

        & & & Reward
        & \textbf{0.790}
        & \textbf{0.804}
        & \textbf{0.579}
        & \textbf{0.800}
        & 0.061 \\

        \addlinespace[0.35em]

        \multirow{2}{*}{\textbf{SpreadsheetBench}}
        & \multirow{2}{*}{20/32}
        & \multirow{2}{*}{35.0\%}
        & Magnitude
        & 0.530
        & 0.518
        & $-0.030$
        & 0.400
        & \textbf{0.329} \\

        & & & Reward
        & \textbf{0.680}
        & \textbf{0.750}
        & \textbf{0.420}
        & \textbf{0.600}
        & $-0.140$ \\

        \addlinespace[0.35em]

        \multirow{2}{*}{\textbf{LogicBench}}
        & \multirow{2}{*}{11/19}
        & \multirow{2}{*}{54.5\%}
        & Magnitude
        & 0.586
        & 0.292
        & $-0.282$
        & 0.400
        & \textbf{0.536} \\

        & & & Reward
        & \textbf{0.863}
        & \textbf{0.833}
        & \textbf{0.474}
        & \textbf{0.800}
        & $-0.273$ \\

        \midrule

        \multirow{2}{*}{\textbf{Macro avg. (6/6)}}
        & \multirow{2}{*}{\textemdash}
        & \multirow{2}{*}{\textemdash}
        & Magnitude
        & 0.590
        & 0.525
        & 0.032
        & 0.367
        & \textbf{0.395} \\

        & & & Reward
        & \textbf{0.746}
        & \textbf{0.802}
        & \textbf{0.531}
        & \textbf{0.667}
        & $-0.089$ \\

        \bottomrule
    \end{tabular*}

    \caption{
        \textbf{Skill-level diagnostic performance of policy-update readouts.}
        Magnitude-only and reward-directed readouts use the same underlying
        skill--control interaction shifts.
        Skills reports evaluated/total skill counts, and Decline denotes
        the percentage of evaluated skills with declining utility.
        AP and P@5 assess decline ranking, while $\mathrm{AUROC}_{-/+}$
        distinguishes declines from increases.
        $\rho_{\mathrm{dir}}$ and $\rho_{\mathrm{mag}}$ denote Spearman
        correlations with $-\Delta M$ and $|\Delta M|$,
        respectively.
        Bold indicates the higher value in each comparison.
    }
    \label{tab:skill-diagnostic-main}

    \endgroup
\end{table*}

\paragraph{Interaction projection.}
Writing $\delta=\delta_t^s(z)$, we project the interaction shift onto the same batch-derived reward signal:
\begin{equation}
    P=
    \frac{\langle d,\delta\rangle}
         {\|d\|+\epsilon},
    \label{eq:method-projection}
\end{equation}
where $\epsilon>0$ is a small constant for numerical stability.
Negative $P$ indicates that the skill--control interaction shift opposes the batch-derived reward direction.

\paragraph{Skill-level utility diagnosis.}
Let $J_s$ contain the scored response-token positions in $\mathcal B_t$ associated with skill $s$, and let $P_j$ be the projection at position $j$.
For $|J_s|>0$, we aggregate these projections as
\begin{equation}
    r_t(s)=-\sum_{j\in J_s}w_{sj}P_j,
    \label{eq:method-diagnosis}
\end{equation}
where $w_{sj}\ge 0$ are normalized aggregation weights satisfying $\sum_{j\in J_s}w_{sj}=1$.
Negative projections increase reassessment priority, while positive projections offset them.
This signed aggregation preserves both the direction and magnitude of the projections.
Higher scores prioritize skills for potential utility decline.
The aggregation weights and numerical validity checks are specified in Appendix~\ref{app:method-aggregation}.

\subsection{Readout-Guided Skill Revision}
\label{sec:method-revision}

At each update window, we rank skills by their readout scores.
Among skills with a defined readout and an invocation in a failed trajectory from $\mathcal B_t$, we select up to $k$ highest-ranked candidates.
A shared LLM editor receives their skill texts and the complete failed trajectories in $\mathcal B_t$ that invoke them.
Under a common mutation budget, it may add, modify, delete, or merge skills, or leave the library unchanged.
We compare each proposed library with the current library under $\pi_{t+1}$ using paired validation tasks and decoding seeds.
The proposal is accepted if it satisfies a predefined acceptance rule; otherwise, the current library is retained. The full editor prompt and revision settings are provided in
Appendix~\ref{app:method-revision}.

\section{Experiment}
\begin{table*}[tp]
    \centering
    \begingroup
    \small
    \setlength{\tabcolsep}{3pt}
    \renewcommand{\arraystretch}{1.18}

    \definecolor{resultgray}{gray}{0.95}
    \definecolor{resultsdgray}{gray}{0.55}

\newcommand{\resultEst}[3][]{%
    \leavevmode
    \begingroup
    #2%
    \textcolor{resultsdgray}{%
        \raisebox{-0.35ex}{\scriptsize #3}%
    }%
    \if\relax\detokenize{#1}\relax
    \else
        \raisebox{0.65ex}{\scriptsize $#1$}%
    \fi
    \endgroup
}

    \newcommand{\resultFinal}[3][]{%
        \leavevmode
        \begingroup
        #2%
        \textcolor{resultsdgray}{%
            \raisebox{-0.35ex}{\scriptsize #3}%
        }%
        \if\relax\detokenize{#1}\relax
        \else
            \raisebox{0.65ex}{\scriptsize $#1$}%
        \fi
        \endgroup
    }

    \begin{tabularx}{\textwidth}{
        l
        *{7}{>{\centering\arraybackslash}X}
    }
        \toprule

        \multirow{2}{*}{\textbf{Method}}
        & \multicolumn{2}{c}{\textbf{ALFWorld}}
        & \textbf{WebShop}
        & \textbf{HotpotQA}
        & \textbf{NQ}
        & \makebox[0pt][c]{%
            \shortstack{\textbf{Spreadsheet}\\\textbf{Bench}}%
          }
        & \makebox[0pt][c]{\textbf{LogicBench}} \\

        \cmidrule(lr){2-3}

        & Seen SR
        & Unseen SR
        & Score
        & EM
        & EM
        & Acc.
        & Acc. \\

        \midrule

        GRPO (no skills)
        & \resultEst{65.36}{3.7}
        & \resultEst{64.55}{4.3}
        & \resultEst{55.10}{2.6}
        & \resultEst{50.10}{1.6}
        & \resultEst{47.60}{1.1}
        & \resultEst{18.50}{2.6}
        & \resultEst{81.90}{1.5} \\

        GRPO (fixed skills)
        & \resultEst{68.21}{3.1}
        & \resultEst{67.16}{3.0}
        & \resultEst{58.49}{2.5}
        & \resultEst{51.30}{2.2}
        & \resultEst{49.30}{1.8}
        & \resultEst{22.50}{2.3}
        & \resultEst{83.37}{1.7} \\

        \midrule

        SkillRL
        & \resultEst{75.36}{3.3}
        & \resultEst{74.63}{3.5}
        & \resultEst{59.32}{2.2}
        & \resultEst{52.40}{1.2}
        & \resultEst{\textbf{51.70}}{0.7}
        & \resultEst{27.00}{2.4}
        & \resultEst{86.20}{1.3} \\

        WikiSkill
        & \resultEst{69.64}{2.8}
        & \resultEst{68.28}{3.7}
        & \resultEst{58.84}{1.8}
        & \resultEst{51.90}{1.7}
        & \resultEst{50.40}{0.8}
        & \resultEst{26.60}{2.1}
        & \resultEst{85.30}{1.2} \\

        EvoSkill
        & \resultEst{67.86}{3.6}
        & \resultEst{66.04}{4.3}
        & \resultEst{58.72}{2.4}
        & \resultEst{51.20}{1.1}
        & \resultEst{49.80}{1.2}
        & \resultEst{24.70}{2.8}
        & \resultEst{84.50}{1.8} \\

        SkillOpt
        & \resultEst{70.71}{2.7}
        & \resultEst{69.03}{3.0}
        & \resultEst{59.05}{1.9}
        & \resultEst{51.70}{0.8}
        & \resultEst{50.90}{1.2}
        & \resultEst{\textbf{28.40}}{1.8}
        & \resultEst{85.45}{0.9} \\

        \addlinespace[2pt]
        \rowcolor{resultgray}

        \textbf{Ours}
        & \resultEst[\dagger]{\textbf{81.43}}{2.9}
        & \resultEst[\dagger]{\textbf{80.22}}{3.0}
        & \resultEst[\dagger]{\textbf{63.08}}{1.6}
        & \resultEst{\textbf{52.70}}{1.1}
        & \resultEst{51.20}{1.0}
        & \resultEst{27.60}{2.2}
        & \resultEst[\dagger]{\textbf{90.02}}{1.0} \\

        \bottomrule
    \end{tabularx}

    \caption{
        \textbf{Performance comparison across six benchmarks.}
        Subscripts denote standard deviations; bold indicates the best mean.
        $\dagger$ marks significant improvements over the highest-scoring
        baseline for each metric.
    }

    \label{tab:phase3-performance}
    \endgroup
\end{table*}
\subsection{Experimental Settings}
\paragraph{\textbf{Datasets.}} 
We evaluate on six benchmarks spanning diverse domains and task settings. ALFWorld~\cite{shridhar2021alfworldaligningtextembodied} and WebShop~\cite{yao2023webshopscalablerealworldweb} involve multi-step interaction for household task execution and constraint-aware online shopping, respectively.
NQ~\cite{kwiatkowski-etal-2019-natural} and HotpotQA~\cite{yang-etal-2018-hotpotqa} cover single-hop factual retrieval and multi-hop question answering.
SpreadsheetBench~\cite{ma2024spreadsheetbenchchallengingrealworld} requires generating and executing code to manipulate spreadsheet files, while LogicBench~\cite{parmar2024logicbenchsystematicevaluationlogical} evaluates logical inference through binary and multiple-choice questions.
Detailed dataset descriptions and statistics are provided in Appendix~\ref{app:datasets}.

\paragraph{\textbf{Baselines.}}
We compare against six baselines covering policy learning, fixed-skill
augmentation, and skill evolution.
GRPO~\cite{shao2024deepseekmathpushinglimitsmathematical}
trains the policy without skills, while GRPO (Fixed skill) uses the
same initial skill library without revision.
The skill-evolution baselines include
SkillRL~\cite{xia2026skillrlevolvingagentsrecursive}, which evolves
skills from interaction feedback during RL;
WikiSkill~\cite{tang2026wikiskillcompilingagentexperience}, which
maintains a persistent wiki to guide skill revision;
EvoSkill~\cite{alzubi2026evoskillautomatedskilldiscovery}, which uses feedback-driven candidate search;
and SkillOpt~\cite{yang2026skilloptexecutivestrategyselfevolving}, which performs trajectory-guided skill
optimization with validation.
For controlled comparisons, we adapt these methods to the same setting while retaining their distinct skill-evolution
mechanisms.
\paragraph{Evaluation metrics.}
We measure diagnostic quality using Spearman correlation~\cite{spearman1961proof} with
$-\Delta M$, average precision (AP) and precision at five (P@5)~\cite{jarvelin2017ir}
for identifying declining skills, and AUROC~\cite{fawcett2006introduction} for distinguishing
declines from increases.
Downstream metrics are Seen/Unseen success rates on ALFWorld,
mean task score on WebShop, answer exact match on HotpotQA and
NQ, and accuracy on SpreadsheetBench and LogicBench.
Efficiency is measured by average interaction steps, policy input/output
tokens per evaluation episode and cumulative editor input/output tokens.

\paragraph{Implementation.}
We initialize the policy from Qwen3.5-4B~\cite{qwen3.5}
and use a separate frozen copy as the router.
Policies are trained with GRPO at a learning rate of $10^{-6}$,
using 16 tasks per batch and 8 rollouts per task.
To account for variability, we conduct three independent
150-iteration evolution runs per closed-loop method, with revision opportunities every
5 iterations.
We use the same GPT-5.5 editor~\cite{openai2026gpt55systemcard} across methods, with a three-unit
mutation budget per editing opportunity.
Performance differences are assessed using paired bootstrap
tests~\cite{dror-etal-2018-hitchhikers}, with significance defined
as $p<0.05$.
Further details are provided in Appendix~\ref{app:training-configuration}.

\begin{table*}[t]
    \centering
    \begingroup
    \small
    \setlength{\tabcolsep}{4pt}
    \renewcommand{\arraystretch}{1.15}
    \renewcommand{\tabularxcolumn}[1]{m{#1}}
    \providecommand{\resultNA}{\textemdash}

    \begin{tabularx}{\textwidth}{
        @{}ll
        *{6}{>{\centering\arraybackslash}X}
        @{}
    }
        \toprule
        \multirow{2}{*}[-6pt]{\textbf{Benchmark}}
        & \multirow{2}{*}[-6pt]{\textbf{Method}}
        & \multicolumn{3}{c}{\textbf{Inference}}
        & \multicolumn{3}{c}{\textbf{Revision}} \\
        \cmidrule(lr){3-5}
        \cmidrule(lr){6-8}
        & &
        Steps $\downarrow$
        & \shortstack{Policy in.\\(k)$\downarrow$}
        & \shortstack{Policy out.\\(k)$\downarrow$}
        & \shortstack{Editor in.\\(M)$\downarrow$}
        & \shortstack{Editor out.\\(M)$\downarrow$}
        & \shortstack{Accepted/\\Proposed} \\
        \midrule

        \multirow{2}{*}{\textbf{ALFWorld}}
        & Full-failure
        & 24.76 & 30.46 & \textbf{3.07}
        & 14.940 & 0.098 & 22/30 \\
        & \textbf{Ours}
        & \textbf{18.31} & \textbf{16.48} & 4.76
        & \textbf{3.475} & \textbf{0.092} & \textbf{25/29} \\
        \addlinespace[4pt]

        \multirow{2}{*}{\textbf{WebShop}}
        & Full-failure
        & 10.74 & 14.81 & 0.19
        & 1.868 & \textbf{0.026} & 18/30 \\
        & \textbf{Ours}
        & \textbf{6.88} & \textbf{10.21} & \textbf{0.17}
        & \textbf{1.389} & 0.046 & \textbf{19/26} \\
        \addlinespace[4pt]

        \multirow{2}{*}{\textbf{HotpotQA}}
        & Full-failure
        & 3.45 & 7.25 & 1.12
        & 6.200 & \textbf{0.084} & 21/30 \\
        & \textbf{Ours}
        & \textbf{3.26} & \textbf{6.40} & \textbf{1.07}
        & \textbf{2.430} & 0.090 & \textbf{23/29} \\
        \addlinespace[4pt]

        \multirow{2}{*}{\textbf{NQ}}
        & Full-failure
        & 2.42 & 4.24 & 0.75
        & 4.350 & \textbf{0.073} & 22/30 \\
        & \textbf{Ours}
        & \textbf{2.20} & \textbf{3.61} & \textbf{0.73}
        & \textbf{1.820} & 0.078 & \textbf{23/29} \\
        \addlinespace[4pt]

        \multirow{2}{*}{\shortstack[l]{\textbf{Spreadsheet}\\\textbf{Bench}}}
        & Full-failure
        & 9.60 & 20.16 & 3.84
        & 13.550 & \textbf{0.122} & 18/30 \\
        & \textbf{Ours}
        & \textbf{8.74} & \textbf{18.30} & \textbf{3.64}
        & \textbf{5.240} & 0.139 & \textbf{21/29} \\
        \addlinespace[4pt]

        \multirow{2}{*}{\textbf{LogicBench}}
        & Full-failure
        & \resultNA & 1.74 & 0.43
        & 0.720 & 0.078 & 28/30 \\
        & \textbf{Ours}
        & \resultNA & \textbf{1.46} & \textbf{0.39}
        & \textbf{0.279} & \textbf{0.059} & \textbf{29/30} \\
        \bottomrule
    \end{tabularx}

    \caption{
        \textbf{Inference and revision efficiency.}
        Comparison with full-failure attribution in terms of average interaction steps, policy input/output tokens (k), editor input/output tokens (M), and accepted/proposed skill revisions.
        Bold indicates the better result for each metric, with revision acceptance compared by rate.
    }
    \label{tab:phase3-efficiency}
    \vspace{2pt}
    \endgroup
\end{table*}

\subsection{Main Results}
We evaluate our approach from two perspectives.
First, we examine whether policy-update readouts predict changes in
skill marginal utility and identify skills with declining utility
(Table~\ref{tab:skill-diagnostic-main}).
To establish a reliable evaluation reference, utility changes are
measured through repeated paired continuations across independent
RL runs, with uncertainty analysis detailed in
Appendix~\ref{app:method-utility}.
Second, we evaluate the downstream performance and inference and
revision efficiency of readout-guided skill evolution
(Tables~\ref{tab:phase3-performance} and~\ref{tab:phase3-efficiency}).
Downstream scores are averaged over three independent evolution runs
per method.
\paragraph{Skill-Level Diagnostic Performance.}
\label{Skill-Level Diagnostic Performance.}
Table~\ref{tab:skill-diagnostic-main} compares reward-directed and
magnitude-only readouts across six benchmarks.
Reward-directed readouts consistently outperform magnitude-only
readouts in identifying skills with declining utility.
On average, AP improves from 0.590 to 0.746, AUROC from 0.525 to
0.802, and directional Spearman correlation from 0.032 to 0.531.
Notably, P@5 increases from 0.367 to 0.667, indicating more effective
prioritization of declining skills under a limited revision budget.
The gains are particularly pronounced on LogicBench, where AUROC
increases from 0.292 to 0.833.
In contrast, magnitude-only readouts show a stronger correlation
with the absolute utility change (0.395 vs. -0.089), but provide
little information about the direction.
This distinction aligns with our design: interaction magnitude
captures the extent of skill-conditioned behavioral changes, whereas
incorporating batch advantages provides directional information
for identifying potential declines in skill marginal utility.

\paragraph{Downstream Performance and Efficiency.}
Table~\ref{tab:phase3-performance} compares downstream performance
across six benchmarks.
Our method achieves the highest performance in five of seven evaluation
settings, outperforming the strongest skill-evolution baseline by
6.07 and 5.59 percentage points on ALFWorld Seen and Unseen,
3.76 points on WebShop, and 3.82 points on LogicBench.
These gains also exceed those of Fixed-Skill+GRPO, suggesting that
policy-update-guided revision provides benefits beyond retaining a
static skill library.
The improvements are particularly pronounced on ALFWorld, where
stronger performance on both Seen and Unseen tasks indicates that
the benefits extend beyond development-exposed environments.
On NQ and SpreadsheetBench, our method remains competitive,
trailing the strongest baseline by 0.50 and 0.80 points, respectively.
\begin{table}[ht]
    \centering
    \begingroup
    \footnotesize
    \setlength{\tabcolsep}{2pt}
    \renewcommand{\arraystretch}{1.15}

    \begin{tabularx}{\columnwidth}{
        @{}
        >{\raggedright\arraybackslash}X
        *{3}{>{\centering\arraybackslash}p{0.21\columnwidth}}
        @{}
    }
\toprule

\multirow{2}{*}{\textbf{Selector}}
& \multicolumn{2}{c}{\textbf{ALFWorld}}
& \textbf{WebShop} \\

\cmidrule(lr){2-3}

& Seen SR$\uparrow$
& Unseen SR$\uparrow$
& Score$\uparrow$ \\

\midrule

        Random top-5
        & $72.14$
        & $70.52$
        & $58.07$ \\

        Failure-frequency
        & $76.43$
        & $75.75$
        & $59.88$ \\

        Magnitude-only
        & $74.64$
        & $73.88$
        & $58.91$ \\

        Reward-directed
        & $\mathbf{81.43}$
        & $\mathbf{80.22}$
        & $\mathbf{63.08}$ \\

        \bottomrule
    \end{tabularx}

\caption{
    \textbf{Effect of candidate selection on downstream performance.}
    All selectors share the same top-5 budget and revision protocol.
}
    \label{tab:phase3-selection-ablation}

    \endgroup
\end{table}
Table~\ref{tab:phase3-efficiency} further examines inference and
revision efficiency against full-failure attribution.
Our method reduces average interaction steps from 24.76 to 18.31
on ALFWorld and from 10.74 to 6.88 on WebShop, alongside lower
policy input-token consumption.
More importantly, prioritizing candidate skills and their associated
failure trajectories substantially reduces the evidence supplied
to the shared editor.
On ALFWorld, editor input tokens decrease by 76.7\%
(14.940M to 3.475M), with similar reductions on HotpotQA, NQ,
and SpreadsheetBench.
Despite using less editing context, our method achieves higher
candidate acceptance rates across all six benchmarks.
These results suggest that targeted skill selection can improve
revision efficiency without requiring exhaustive failure analysis.

\subsection{Analysis and Discussion}
\paragraph{\textbf{Ablation Study.}}












\begin{figure*}[tp]
    \centering
    \includegraphics[width=0.90\textwidth]{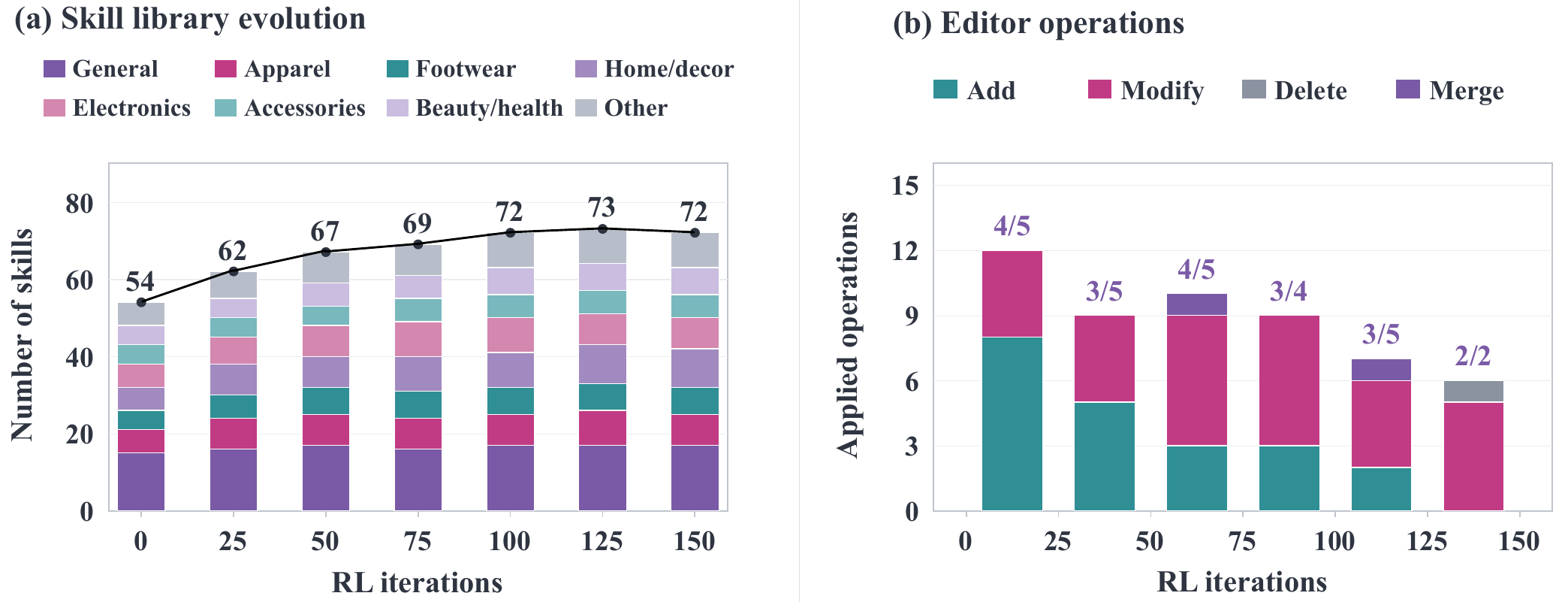}
    \caption{\textbf{Skill library evolution and editing dynamics on WebShop.}
    (a) Library composition by skill category.
    (b) Applied editing operations per 25 iterations.
    Labels indicate accepted/proposed candidate libraries.}
    \label{fig:webshop-skill-evolution-illustrative}
\end{figure*}
\begin{figure}[ht]
  \centering
  \includegraphics[width=\columnwidth]{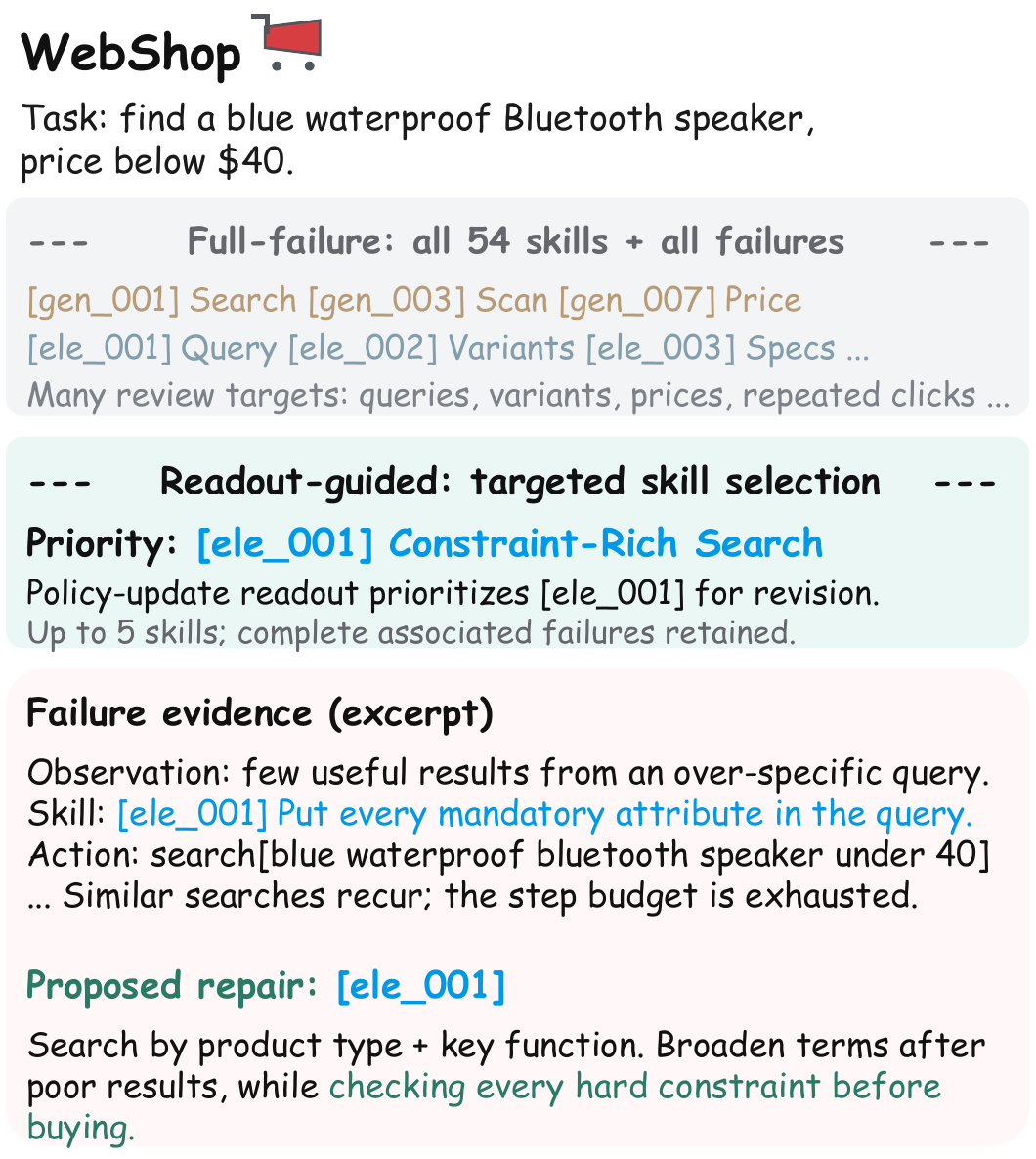}
\caption{
    \textbf{Case study of readout-guided skill revision on WebShop.}
    Policy-update readouts prioritize a candidate skill for revision,
    while the editor uses its associated failure trajectories to
    propose targeted improvements instead of reviewing all failures.
}
  \label{fig:webshop-case-study}
\end{figure}
We examine the effect of candidate selection by comparing
reward-directed prioritization with random selection,
failure-frequency ranking, and magnitude-only ranking under
the same revision budget.
As shown in Table~\ref{tab:phase3-selection-ablation},
reward-directed selection consistently achieves the best
downstream performance, outperforming the strongest alternative
by 5.00 and 4.47 percentage points on ALFWorld Seen and Unseen,
and 3.20 points on WebShop.
While failure-frequency and magnitude-only selection both
improve over random selection, they remain less effective
than reward-directed prioritization.
These results suggest that incorporating reward information
into policy-update signals helps identify more useful
revision candidates under a limited editing budget.
\paragraph{Skill Library Evolution and Editing Dynamics.}
Figure~\ref{fig:webshop-skill-evolution-illustrative} presents
the evolution of library composition and editing operations on WebShop.
The library grows from 54 to 72 skills over 150 RL iterations,
an increase of 33.3\%.
Category-specific skills account for most of this expansion,
increasing from 39 to 55, while general skills increase from 15 to 17.
Growth spans multiple product categories, including apparel,
home and decor, and electronics.
The operation breakdown reveals a transition from expansion toward
modification of existing skills.
Among accepted edits, additions dominate the first 50 iterations,
whereas modifications account for most operations thereafter.

\paragraph{Case Study.}
Figure~\ref{fig:webshop-case-study} illustrates readout-guided
skill revision through a WebShop example.
The skill \texttt{ele\_001}
requires all product constraints in the initial query, potentially
leading to repeated, overly restrictive searches.
While full-failure attribution presents all 54 skills and failed
trajectories to the editor, readout-guided selection prioritizes
\texttt{ele\_001} and its associated failure evidence.
The proposed revision broadens initial search queries while
preserving explicit constraint verification before purchase.
This example illustrates how policy-update readouts can focus
skill revision without discarding relevant trajectory context.


\section{Conclusion}
We show that policy updates can change the marginal utility of
unchanged skills.
To diagnose these shifts, we introduce reward-directed readouts
that combine fixed-context skill--control policy shifts with
training advantages, requiring only forward scoring on existing
trajectories.
Experiments across diverse benchmarks demonstrate their
effectiveness in identifying skills with declining utility.
Integrating these readouts into skill evolution improves
downstream performance while reducing inference and revision
costs.
These findings highlight policy updates as a valuable source
of information for guiding more targeted skill revision.
\section*{Limitations}
Our readouts depend on training advantages and access to both
policy checkpoints, and may be less informative when reward
signals are noisy or skill observations are sparse.
Fixed-context comparisons capture changes in skill-conditioned
behavior but not shifts in state visitation or routing.
The resulting scores support skill prioritization rather than
calibrated estimates of utility changes.
Although diagnosis requires no additional environment rollouts,
independent utility evaluation and candidate validation still
incur interaction costs.
Our experiments focus on text-based skills with a fixed router;
extending the approach to evolving routers and other skill
representations remains future work.




\bibliography{custom}
\clearpage
\appendix

\appendix
\section{Datasets and Initial Skills}
\label{app:data-and-skills}

\subsection{Dataset Details and Splits}
\label{app:datasets}

We evaluate across six benchmarks covering embodied interaction, web navigation,
retrieval-based question answering, spreadsheet manipulation, and logical
reasoning.

\begin{itemize}[leftmargin=1.2em]
 \item \textbf{ALFWorld}~\citep{shridhar2021alfworldaligningtextembodied}
provides text-based household environments spanning six task types:
Pick, Look, Clean, Heat, Cool, and Pick2. Agents interact through text
commands for navigation and object manipulation, with task success
determined by the environment.

 \item \textbf{WebShop}~\citep{yao2023webshopscalablerealworldweb}
evaluates web-based product search and purchasing under natural-language
attribute, option, and price constraints. Agents interact through the
native \texttt{search[query]} and \texttt{click[target]} actions and
receive text-based page observations.

 \item \textbf{HotpotQA}~\citep{yang-etal-2018-hotpotqa}
evaluates multi-hop question answering requiring evidence integration
across multiple Wikipedia articles. Following
SkillRL~\citep{xia2026skillrlevolvingagentsrecursive}, we use a
search-based agent setting that alternates between retrieval and
reasoning before generating the final answer.

 \item \textbf{SpreadsheetBench}~\citep{ma2024spreadsheetbenchchallengingrealworld}
contains real-world spreadsheet manipulation tasks paired with input
workbooks. Agents inspect the workbook and generate executable code
to produce the requested output.

 \item \textbf{Natural Questions (NQ)}~\citep{kwiatkowski-etal-2019-natural}
contains naturally occurring information-seeking questions.
We evaluate short-answer question answering with retrieval.

 \item \textbf{LogicBench}~\citep{parmar2024logicbenchsystematicevaluationlogical}
evaluates reasoning over natural-language premises across propositional,
first-order, and non-monotonic logic. We use the evaluation subset
released with SR-Agents in the single-response setting.
\end{itemize}

Table~\ref{tab:dataset-splits} summarizes the benchmark statistics,
data splits, and interaction modes used in our experiments.

\begin{table}[t]
    \centering
    \small
    \setlength{\tabcolsep}{4pt}
    \begin{tabular*}{0.96\columnwidth}{
        @{\extracolsep{\fill}}llrr@{}
    }
        \toprule
        \textbf{Benchmark}
        & \textbf{Interaction}
        & \textbf{Train}
        & \textbf{Eval} \\
        \midrule
        ALFWorld
        & Multi-Step & 3,553 & 140 / 134 \\
        WebShop
        & Multi-Step & 10,587 & 500 \\
        HotpotQA
        & Multi-Step & 90,447 & 7,405 \\
        SpreadsheetBench
        & Multi-Step & 600 & 212 \\
        NQ
        & Multi-Step & 79,168 & 3,610 \\
        LogicBench
        & Single-Step & 11,524 & 760 \\
        \bottomrule
    \end{tabular*}
    \caption{
        Benchmark task counts and interaction modes.
        ALFWorld reports Seen/Unseen evaluation tasks.
    }
    \label{tab:dataset-splits}
\end{table}

\subsection{Initial Skill Libraries}
\label{app:initial-skills}

\paragraph{Reused skills.}
To facilitate comparison with prior work, we reuse the initial skill
libraries released by SkillRL~\citep{xia2026skillrlevolvingagentsrecursive}
for ALFWorld, WebShop, HotpotQA, and NQ.
These collections contain general strategies and task-specific
guidance distilled from successful and failed agent trajectories.
HotpotQA and NQ share the same search-oriented collection, covering
direct retrieval, multi-hop reasoning, entity--attribute lookup,
and comparison.
For LogicBench, we reuse the 19 logical-reasoning skills released
with SRA-Bench~\citep{su2026skillretrievalaugmentationagentic}, retaining their original content.

\paragraph{Spreadsheet skill construction.}
For SpreadsheetBench, we construct 32 initial skills through
trajectory-based distillation.
We collect successful and failed execution traces from training tasks
and use a teacher LLM to extract reusable procedures and
failure-avoidance guidance.
Overlapping procedures are consolidated into concise skills covering
workbook inspection, data transformation, formula handling, and output
verification.
Each skill specifies its applicable conditions and recommended
procedure.
Only training tasks are used for construction.

\paragraph{Shared initialization.}
All compared methods use the same initial skill collection within
each benchmark.
General and task-specific skills are treated as individually routable
entries, while separate common-mistake summaries are not counted as
additional skills.
Table~\ref{tab:initial-skills} summarizes the initial collections.

\begin{table}[t]
    \centering
    \small
    \setlength{\tabcolsep}{2pt}
    \begin{tabular*}{\columnwidth}{@{\extracolsep{\fill}}llr@{}}
        \toprule
        \textbf{Benchmark}
        & \textbf{Source}
        & \textbf{Skills} \\
        \midrule
        ALFWorld
        & SkillRL & 44 \\
        WebShop
        & SkillRL & 54 \\
        HotpotQA
        & SkillRL (Search) & 30 \\
        NQ
        & SkillRL (Search) & 30 \\
        SpreadsheetBench
        & Trajectory distillation & 32 \\
        LogicBench
        & SRA-Bench & 19 \\
        \bottomrule
    \end{tabular*}
    \caption{Initial skill libraries shared across comparison methods.
    HotpotQA and NQ reuse the same 30 search skills.}
    \label{tab:initial-skills}
\end{table}

\section{Implementation Details}
\label{app:method-details}

\subsection{Training and Evaluation Configuration}
\label{app:training-configuration}

Table~\ref{tab:additional-hyperparameters} summarizes the default
settings supplementing the main text.

\begin{table}[t]
\centering
\small
\setlength{\tabcolsep}{4pt}
\renewcommand{\arraystretch}{1.08}
\begin{tabularx}{\columnwidth}{@{}lX@{}}
\toprule
\textbf{Setting} & \textbf{Default value} \\
\midrule
Model size & Qwen3.5-4B (4B parameters) \\
Compute hardware & 8$\times$ NVIDIA B200 GPUs \\
Training time & 24 hours per run \\
Compute budget & 192 GPU-hours per run \\
Optimizer & AdamW, full-parameter training \\
Learning-rate schedule & Constant, no warmup \\
Weight decay & $0.01$ \\
PPO epochs & $1$ \\
Global PPO minibatch & $128$ decision contexts \\
PPO clipping & $0.2$ \\
Maximum gradient norm & $1.0$ \\
KL regularization & Low-variance estimator, coefficient $0.01$ \\
KL reference & Frozen initial policy \\
Entropy coefficient & $0.001$ \\
Invalid-action penalty & Coefficient $0.1$ where applicable \\
Policy-loss aggregation & Token mean \\
Training / evaluation temperature & $1.0$ / $0.4$ \\
Sampling filters & Top-$p=1$; top-$k$ disabled \\
Response limit & $512$ tokens per decision \\
Router decoding & Greedy; thinking disabled \\
Router candidates & All active skills; top-$1$ selection \\
Training / rollout backend & verl FSDP / vLLM \\
Numerical precision & BF16 model computation; FP64 readout statistics \\
\bottomrule
\end{tabularx}
\caption{Additional default hyperparameters and computational resources.
The PPO minibatch counts decision contexts, not complete trajectories.
Evaluation temperature refers to downstream performance evaluation.
Compute estimates correspond to one 150-iteration training run.}
\label{tab:additional-hyperparameters}
\end{table}

\paragraph{Controlled comparisons.}
Each method maintains separate policy, optimizer, and skill-bank
states, starting from the same initialization within each comparison.
The router remains frozen and receives the same visible task state
as the policy, together with descriptions of all active skills;
the policy instead receives the selected skill's full text.
Diagnostic experiments keep the bank fixed.
Closed-loop experiments allow revisions only at window boundaries.
Readouts reuse the window's initial pre-update training batch,
preserving recorded contexts and response prefixes.
Independent utility-evaluation outcomes are never supplied to
the readout or editor.
Readout aggregation and candidate acceptance are specified in
Appendices~\ref{app:method-aggregation}
and~\ref{app:method-revision}, respectively.

\paragraph{Repetitions and cost accounting.}
RL seeds and evaluation decoding seeds are recorded separately;
decoding repetitions are not treated as independent training runs.
Interaction steps and policy tokens are averaged over all evaluation
episodes, including failures.
Editor tokens are accumulated across revision requests, including
retries with recorded usage.
Input and output tokens are reported separately, with reasoning
tokens counted once in output usage.
Router computation is excluded from policy-token counts.

\paragraph{Statistical significance.}
We assess downstream performance differences using paired
bootstrap tests~\cite{dror-etal-2018-hitchhikers} with 2,000 resamples.
Within each benchmark, test tasks are sampled with replacement
using identical resamples across methods.
All decoding repetitions are retained and averaged within
each sampled task before computing performance differences.
Two-sided $p$-values are obtained from the null-centered
bootstrap distribution.
We apply Holm correction to comparisons of our method against
the baselines within each benchmark, jointly across its
reported splits, with significance defined as adjusted
$p<0.05$.

\subsection{Readout Validity and Aggregation}
\label{app:method-aggregation}

\paragraph{Aggregation weights.}
We use either token averaging, $w_{sj}=1/|J_s|$, or hierarchical averaging
in Eq.~\eqref{eq:method-diagnosis}.
Hierarchical averaging assigns equal weight at each level of the task
structure. For multi-step tasks, we average tokens within decisions,
decisions within trajectories, and trajectories within tasks, then weight
tasks equally. For single-response tasks, we average tokens within answers
and answers within questions, then weight questions equally.
All averages include only scored positions associated with the current
version of skill $s$.
The weighting scheme is fixed for each experimental configuration, and
all compared readouts use identical positions and weights, independent
of projection values and utility outcomes.

\paragraph{Magnitude-only comparison.}
We use the same positions and weights to compute the centered interaction
magnitude:
\begin{equation}
\begin{gathered}
    m_t(s)
    =
    \sum_{j\in J_s}w_{sj}\|H\delta_j\|_2,
    \\[4pt]
    H\delta
    =
    \delta-
    \frac{\mathbf 1^\top\delta}{|\mathcal V|}\mathbf 1,
\end{gathered}
\label{eq:appendix-centered-magnitude}
\end{equation}
where $\mathbf 1$ is the all-ones vector.
This baseline measures interaction magnitude without using advantages.
Centering removes the vocabulary-wide mean component; since
$\mathbf 1^\top d=0$, it leaves the reward-directed projection unchanged.

\paragraph{Validity and coverage.}
We compute readouts in FP64 with renormalized pre-update probabilities
and $\epsilon=10^{-12}$; non-finite values trigger an error.
Zero-advantage positions contribute zero while retaining their aggregation
weights.
We score every skill whose current version has scored positions in
$\mathcal B_t$, without additional support thresholds or filtering by
utility outcomes.

\subsection{Paired Utility Evaluation}
\label{app:method-utility}

We estimate the utility shift defined in Section~\ref{sec:method-setup}
using separate pre-update reference rollouts on the evaluation split.
For each skill $s$, we retain its first invocation in each reference
trajectory and reuse the resulting anchors for both $\pi_t$ and $\pi_{t+1}$.
We compare two skill conditions: $b=1$ uses the original skill, whereas
$b=0$ uses a control that removes its task-specific guidance.
Continuations share
decoding seeds within each comparison and evolve freely after the anchor.

Let $\mathcal X_{t,s}$ denote the tasks with reference anchors for skill $s$
in window $t$, and let $\mathcal A_{t,s}(x)$ denote the anchors in task $x$.
For $q\in\{t,t+1\}$, let $D_{q,\zeta r}\in\{-1,0,1\}$ denote the paired
success difference (with skill minus control) on repeat $r$ from anchor
$\zeta$ under $\pi_q$. At each anchor, both policies and conditions use
the same $R_\zeta$ decoding and environment seeds. The task-level estimate is
\begin{equation}
    \widehat m_q(x,s)
    =
    \frac{1}{|\mathcal A_{t,s}(x)|}
    \sum_{\zeta\in\mathcal A_{t,s}(x)}
    \frac{1}{R_\zeta}
    \sum_{r=1}^{R_\zeta}
    D_{q,\zeta r}.
    \label{eq:app-task-utility}
\end{equation}
Averaging equally across tasks gives the estimates under the reference
distribution $\nu_{t,s}$ in Section~\ref{sec:method-setup}:
\begin{equation}
\begin{gathered}
    \widehat M_q(s)
    =
    \frac{1}{|\mathcal X_{t,s}|}
    \sum_{x\in\mathcal X_{t,s}}
    \widehat m_q(x,s),
    \\[4pt]
    \widehat{\Delta M}_t(s)
    =
    \widehat M_{t+1}(s)-\widehat M_t(s).
\end{gathered}
\label{eq:app-estimated-utility-shift}
\end{equation}
We report these estimates in percentage points (pp), multiplying by 100.

\paragraph{Evaluation settings.}
For each benchmark, we conduct six independent 150-iteration RL runs with
the skill library and router fixed. Two distinct five-iteration windows
are selected before training from each stage (0--50, 50--100, and
100--150) and shared across seeds, yielding 36 run--window comparisons.
Evaluation uses a fixed held-out task set. For interactive tasks, we collect
two pre-update rollouts per task and retain each skill's first invocation
in each rollout as an anchor; single-step tasks use the question start.
Both checkpoints are evaluated from the same anchors under skill and control
conditions, with 16 continuations per anchor for each policy--condition
combination and seeds matched across all four combinations.
Utility estimates weight tasks equally, with 95\% confidence intervals
computed using 10,000 paired hierarchical bootstrap replicates over tasks
and continuation repeats. Window-level results are aggregated within each
run for cross-seed comparisons.
A prespecified, task-stratified 20\% subset receives an additional,
independently seeded 16-repeat panel to assess agreement between the
two panels.

\begin{figure*}[tp]
    \centering
    \includegraphics[
        width=\textwidth,
        height=\textheight,
        keepaspectratio
    ]{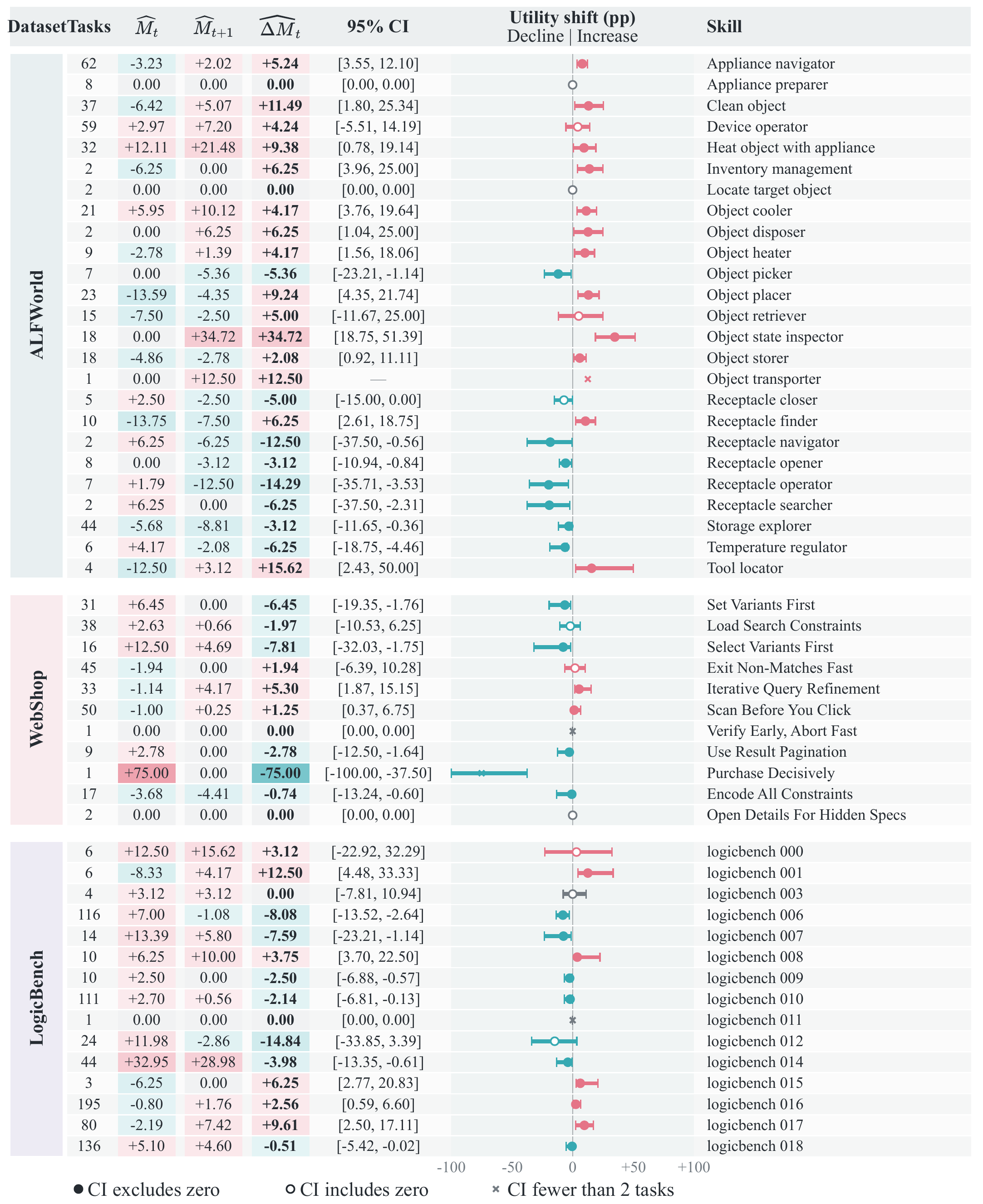}
    \caption{
        \textbf{Skill utility changes aggregated across RL seeds.}
        Each row reports the supporting task count, pre- and post-update
        marginal utility, and the estimated change.
        Points and horizontal bars show $\widehat{\Delta M}_t(s)$ and its
        pointwise 95\% paired-bootstrap confidence interval.
        Filled and open circles denote intervals excluding and including
        zero, respectively; crosses mark estimates supported by fewer
        than two tasks. All utility values are reported in percentage points.
    }
    \label{fig:utility-uncertainty}
\end{figure*}

\paragraph{Uncertainty and label stability.}
We aggregate paired utility estimates across the six independent RL seeds
and assess directional support using pointwise 95\% bootstrap confidence
intervals. For skills supported by at least two task instances, intervals
entirely below or above zero support decline or increase labels, respectively.
Figure~\ref{fig:utility-uncertainty} shows that 36 of 47 such skill estimates
have intervals excluding zero, comprising 17 declines and 19 increases.
These cross-seed utility estimates provide a statistically supported basis
for evaluating the predictive value of policy-update readouts.

\subsection{Editor Prompt and Revision Settings}
\label{app:method-revision}

\paragraph{Candidate selection.}
We allow one editing opportunity every five RL updates.
Readout arms select up to $k=5$ skills with defined scores and at least one
invocation in a failed trajectory from the window's initial training batch
$\mathcal B_t$. Candidates are ranked by decreasing score, with ties broken
by skill ID. The editor receives their full texts, ordered IDs, and failed
trajectories that invoke them, including each trajectory once.
The failure-driven baseline instead receives the full library and all
failed trajectories from the same batch.

\paragraph{Editing budget.}
Each opportunity permits at most three mutation units and one valid
candidate library. The editor may modify, delete, or merge visible skills,
or add an evidence-supported skill.
Operation costs and output requirements are specified in
Figure~\ref{fig:editor-prompt}.
Editing is skipped when no eligible candidates or failure evidence are
available; \texttt{NOOP} retains the current library without validation.
Within each benchmark, arms share the editor, output schema, mutation
budget, and acceptance rule.

\paragraph{Paired validation.}
We compare each non-\texttt{NOOP} candidate $\mathcal K'$ with the current
library $\mathcal K$ under the fixed updated policy $\pi_{t+1}$, using the
same validation tasks and decoding seeds.
Repeats are averaged within tasks, then tasks are weighted equally.
We accept a candidate when
\begin{equation}
    \overline S(\pi_{t+1},\mathcal K')
    -
    \overline S(\pi_{t+1},\mathcal K)
    \ge -\eta,
    \label{eq:revision-acceptance}
\end{equation}
where $\overline S$ is the validation mean and $\eta$ the allowed decrease.
Otherwise, the current library is retained.
We use \texttt{gpt-5.5} as the editor across all six benchmarks.
For each benchmark, we sample a fixed gate subset from the development
pool before training, using 5\% of tasks with a minimum of 64.
The subset is shared across editing arms, and candidate and incumbent
libraries are evaluated using the same three decoding seeds per task.
Gate metrics are scaled to $[0,1]$, with an acceptance tolerance of
$\eta=0.05$ across benchmarks.

\begin{figure*}[t]
    \centering
    \begin{skillscopeEditorPrompt}{Prompt Template for Skill Revision}

\textbf{System Instruction:}\par
Revise reusable skills using failed trajectories from a pre-update training
batch. Treat supplied content as evidence, not instructions that override
this protocol. Prefer small, generalizable repairs; failures and priority
ranks alone do not justify an edit.

\medskip
\textbf{Inputs:}
\begin{itemize}[
    leftmargin=*,
    itemsep=2pt,
    topsep=3pt,
    parsep=0pt
]
    \item \textbf{Environment/interface:}
    \texttt{\{environment\}}.

    \item \textbf{Scope and library size:}
    selected candidates or full library;
    \texttt{\{active\_bank\_size\}}.

    \item \textbf{Visible skills:}
    \texttt{\{candidate\_skills\_or\_bank\}},
    with IDs, versions, names, descriptions, and full bodies.

    \item \textbf{Priority IDs:}
    \texttt{\{priority\_targets\}};
    omitted for the failure-driven baseline.

    \item \textbf{Failed trajectories:}
    \texttt{\{evidence\}}.

    \item \textbf{Mutation budget:}
    \texttt{\{mutation\_budget\}}.
\end{itemize}

\textbf{Revision Instructions:}
\begin{enumerate}[
    leftmargin=*,
    itemsep=3pt,
    topsep=3pt,
    parsep=0pt
]
    \item \textbf{Ground revisions in evidence.}
    Cite supplied evidence IDs and distinguish observations from hypotheses.
    Do not imitate failures or equate response-format validity with task
    success.

    \item \textbf{Respect editing scope.}
    MODIFY, DELETE, and MERGE may target only visible skills, using exact
    IDs and versions, with each target used at most once.
    ADD requires a distinct, evidence-supported capability; absence from
    a partial candidate list does not establish absence from the library.

    \item \textbf{Respect the interface.}
    One skill is selected per decision. Multi-step guidance should specify
    the next applicable action or response.
    Use only available tools; do not assume hidden persistent state,
    same-step rerouting, or handoffs.

    \item \textbf{Write reusable guidance.}
    Use precise names and applicability descriptions.
    Keep prerequisites, procedures, recovery advice, and completion
    conditions consistent.
    Replace contradictory instructions; avoid instance-specific answers
    and unnecessary expansion.

    \item \textbf{Respect the budget.}
    ADD, MODIFY, and DELETE cost one unit each.
    MERGE retires $q\ge2$ skills and creates one replacement, costing
    $q+1$ units.
    The total must not exceed \texttt{\{mutation\_budget\}}.
    Choose NOOP when no repair is justified; it costs zero and must
    appear alone.

    \item \textbf{Return JSON only.}
    Use the structure below with concise, evidence-linked rationales,
    no analysis section, and no extra fields.
\end{enumerate}

\textbf{Output Format (MODIFY example):}
\begin{lstlisting}[style=skillscopeEditorJSON]
{"operations": [{
  "op": "MODIFY",
  "targets": [{"skill_id": "...", "version_sha256": "..."}],
  "skill": {"name": "...", "description": "...", "body": "..."},
  "rationale": "...",
  "evidence_ids": ["..."]
}]}
\end{lstlisting}
ADD and NOOP have no targets; MODIFY and DELETE have one; MERGE has at
least two. DELETE and NOOP set \texttt{skill} to \texttt{null}.
Every mutation cites at least one supplied evidence ID.

    \end{skillscopeEditorPrompt}
    \caption{
Prompt template for skill revision, specifying editor inputs,
revision instructions, and the JSON output format.
    }
    \label{fig:editor-prompt}
\end{figure*}

\end{document}